\pdfoutput=1

\documentclass[11pt]{article}

\usepackage{acl}

\usepackage{times}
\usepackage{latexsym}

\usepackage[T1]{fontenc}

\usepackage[utf8]{inputenc}

\usepackage{microtype}

\usepackage{inconsolata}

\usepackage{graphicx}

\usepackage[explicit]{titlesec}
\usepackage{xcolor}
\definecolor{ForestGreen}{RGB}{34,139,34}
\definecolor{Firebrick}{RGB}{178,34,34}
\definecolor{titlecolor}{RGB}{255, 255, 255}
\definecolor{boxcolor}{RGB}{82, 117, 167}

\usepackage[breakable]{tcolorbox}

\usepackage{subcaption}
\usepackage{array}
\usepackage{booktabs}
\usepackage{amsmath}

\title{REFER: Mitigating Bias in Opinion Summarisation via F\underline{re}quency \underline{F}ram\underline{e}d P\underline{r}ompting}

\author{Nannan Huang \\
  RMIT University, Australia \\
  \texttt{amber.huang@student.rmit.edu.au} 
  \And
  Haytham M. Fayek \\
  RMIT University, Australia \\
  \texttt{haytham.fayek@ieee.org} 
  \AND
  Xiuzhen Zhang \\
  RMIT University, Australia \\
  \texttt{xiuzhen.zhang@rmit.edu.au}}

\begin{document}
\maketitle
\begin{abstract}
Individuals express diverse opinions, a fair summary should represent these viewpoints comprehensively.
Previous research on fairness in opinion summarisation using large language models (LLMs) relied on hyperparameter tuning or providing ground truth distributional information in prompts. However, these methods face practical limitations: end-users rarely modify default model parameters, and accurate distributional information is often unavailable. 
Building upon cognitive science research demonstrating that frequency-based representations reduce systematic biases in human statistical reasoning by making reference classes explicit and reducing cognitive load, this study investigates whether frequency framed prompting (REFER) can similarly enhance fairness in LLM opinion summarisation. Through systematic experimentation with different prompting frameworks, we adapted techniques known to improve human reasoning to elicit more effective information processing in language models compared to abstract probabilistic representations.
Our results demonstrate that REFER enhances fairness in language  models when summarising opinions. 
This effect is particularly pronounced in larger language models and using stronger reasoning instructions.


\end{abstract}

\section{Introduction}
Large Language Models (LLMs) have demonstrated exceptional capabilities in general language tasks~\cite{brown2020language, radford2019language, chowdhery2023palm, touvron2023llama, le2023bloom}. However, their training on uncurated datasets inadvertently incorporates societal biases, leading to the perpetuation of social stereotypes in both the models and their downstream applications~\cite{vig2020causal, sheng2019woman, liang2021towards, gallegos2024bias, li2023survey, huang-etal-2023-examining, feng-etal-2023-pretraining, huang-etal-2024-bias}, influencing how users process and interpret information~\cite{jakesch2023co, durmus2023towards, epstein2023art}. This challenge has motivated research into the cognitive mechanisms of LLMs, such as developing human-inspired debiasing strategies~\cite{madaan2024self, furniturewala2024thinking}.

Previous research has demonstrated that human decision-making often relies on cognitive heuristics rather than factual analysis. When forming judgements, individuals tend to utilise readily accessible information instead of considering empirical evidence, base rates, and statistical distributions~\cite{tversky1973availability}. 
While these evolved shortcuts generally serve us well when solving everyday questions, they can lead to systematic biases when applied to contexts requiring precise statistical reasoning~\cite{tversky1974judgment}.
Approaches to cognitive debiasing such as direct warnings about overconfidence, have demonstrated limited to no effectiveness~\cite{larrick2004debiasing}. In contrast, facilitating the transition from abstract reasoning to frequency-based cognitive frameworks has been shown to improve inference accuracy~\cite{tversky1983extensional, fiedler1988dependence, gigerenzer1994distinction}. This is achieved by reframing abstract probability questions into explicit frequency-based terms (e.g., "how likely is this outcome?" becomes "out of 100 cases, how many would show this outcome?"). This approach improves systematic critical reasoning by providing a concrete numerical framework for analysis~\cite{gigerenzer1994distinction, gigerenzer1995improve, gigerenzer1999overcoming}.

When making predictions, humans often deviate from probability theory, overlooking diverse perspectives and weighting information toward their own biases~\cite{allahverdyan2014opinion, wason1960failure, hart2009feeling}. These cognitive limitations stem from how humans process statistical information: we struggle with abstract probability representations but excel when the same information is presented in concrete frequency formats~\cite{gigerenzer1994distinction, cosmides1996humans}.
Language models exhibit oversimplification patterns and biases similar to human cognitive heuristics~\cite{acerbi2023large, xie2023adaptive}. Similar to how humans benefit from frequency-based representations over abstract probabilities, LLMs trained on human-generated text may have internalised comparable representational preferences~\cite{acerbi2023large, hagendorff2023human}. Consequently, LLMs often produce summaries that inadequately represent the full spectrum of opinions~\cite{dash2019summarizing, huang-etal-2023-examining, zhang2023fair, huang-etal-2024-bias}, exhibiting the same selective attention patterns observed in human reasoning.
Current debiasing approaches focus primarily on hyperparameter optimisation and explicit distribution prompts~\cite{zhang2023fair}. However, these methods face practical limitations: end-users rarely modify default parameters, and accurate distributional information is often unavailable. These constraints motivate exploring frequency-based debiasing strategies, which cognitive science research shows improve human statistical reasoning and could enhance LLM opinion summarisation without requiring user configuration or distributional specifications. This approach remains unexplored for LLM opinion summarisation.

Building on this potential, we hypothesise that frequency framing enhances LLM fairness through analogous computational mechanisms to those observed in humans. Specifically, frequency prompts should: (1) specify explicit reference classes by directing model attention to concrete distributional information rather than abstract concepts, (2) provide numerical anchoring that reduces the computational complexity of proportion estimation during text generation, and (3) enable sequential deliberation by activating statistical reasoning patterns learnt during training on frequency-formatted data. Unlike probability statements, which require normalisation and comparison operations, frequency statements provide explicit numerical anchors that should guide attention allocation during summarisation.
This work explores the potential of frequency framing to elicit reasoning and summarising with frequency-based information, we refer to this framework as F\underline{re}quency \underline{F}ram\underline{e}d P\underline{r}ompting---REFER. 


Our key contributions are:
\begin{itemize}
    \item We propose and evaluate an end-user focused framework---REFER for mitigating bias in summarising opinions using language models by eliciting reasoning with frequency-based information.
    \item We show that REFER improves fairness in opinion summarisation across multiple prompting methods, with strongest gains when combined with reasoning-based approaches such as Chain-of-Thought.
    \item Our experiments evaluate REFER's effectiveness across multiple datasets, models and evaluation metrics, demonstrating particular improvements with larger models that have stronger instruction-following capabilities.

\end{itemize}


\section{Related Work}
\subsection{LLMs Prompting}
Research on LLMs prompting has progressed from Chain-of-Thought reasoning~\cite{wei2022chain} to granular task decomposition methods~\cite{zhou2022least, wang2023plan, yao2024tree, taveekitworachai2024null, weinzierl2024tree} and role-based approaches that leverage specific personas~\cite{pataranutaporn2021ai, wang2023rolellm, gupta2023bias}. Recent work has expanded into multi-agent systems where LLMs collaborate through emergent behaviours and distributed decision-making~\cite{liang2023encouraging, long2024multi, zhang2024chain}. However, the application of these advanced frameworks to bias in opinion summarisation remains understudied.

\subsection{Debiasing Through Model Editing}
Debiasing language models has primarily relied on algorithmic interventions through retraining and fine-tuning on balanced datasets~\cite{liu2019does, saunders2020reducing, zmigrod2019counterfactual, ghanbarzadeh2023gender}. As model sizes grow, these approaches become impractical due to computational demands. Alternative methods such as post-training pruning~\cite{zayed2024fairness, yang2024mitigating} and machine unlearning~\cite{liu2024towards, chen2024fast} aim to mitigate bias by removing or suppressing biased model components. However, these methods require direct model access, limiting their use to open-source models. This creates a need for lightweight debiasing methods that work with proprietary models commonly used by the public.

\subsection{Prompting and Bias Mitigation}
Researchers have explored prompting frameworks to address language model bias, from few-shot prompts~\cite{si2022prompting} and fine-tuning~\cite{borchers2022looking} to automated prompt-search~\cite{ma2023fairness}, perspective-taking~\cite{xu2024walking}, and slower reasoning~\cite{furniturewala2024thinking}. While these approaches can mitigate general bias, they lack mechanisms for preserving opinion distributions in summarisation. For opinion summarisation specifically, prior studies have explored debiasing through hyperparameter tuning and ground-truth distributions in prompts~\cite{zhang2023fair, huang-etal-2024-bias}. 
However, end-users rarely modify default model hyperparameters, and ground-truth labels are generally unavailable, highlighting the need for practical debiasing methods in summarisation.

\newtcolorbox{bluebox}[2][]{
    breakable,  
    colback=white,
    colframe=boxcolor,
    top=0pt,
    bottom=0pt,
    left=6pt,
    right=6pt,
    boxrule=0.5pt,
    bottomrule=0.5pt,
    toptitle=1mm,
    bottomtitle=1mm,
    title={\textcolor{white}{\textsf{\footnotesize #2}}},
    colbacktitle=boxcolor,
    fonttitle=\footnotesize,
    fontupper=\ttfamily\footnotesize\raggedright,
    before skip=8pt,
    after skip=8pt,
    #1
}

\section{F\underline{re}quency \underline{F}ram\underline{e}d P\underline{r}ompting (REFER)}


Research demonstrates that frequency formats facilitate more accurate statistical reasoning and reduce bias compared to abstract probabilities~\cite{cosmides1996humans, hoffrage2000communicating}. Framing questions to elicit frequency-based responses rather than abstract probabilities has been shown to promote critical thinking in human reasoning~\cite{gigerenzer1994distinction, gigerenzer1995improve, gigerenzer1999overcoming}. This improvement occurs through three key mechanisms: (1) specifying explicit reference classes through concrete denominators and comparison groups, (2) providing numerical anchoring that simplifies the representation of uncertainty, and (3) enabling sequential deliberation by aligning with cognitive mechanisms that separate quantitative analysis from intuitive judgement~\cite{sedlmeier2001teaching}. 
Building upon this theoretical foundation, we introduce F\underline{re}quency \underline{F}ram\underline{e}d P\underline{r}ompt (REFER), designed to activate these same computational mechanisms in language models. REFER operationalises the three cognitive science principles as follows: \textbf{Reference Class Specification:} Rather than asking models to generate `balanced' summaries (an abstract instruction), REFER uses frequency-based framing: `determine how many reviews out of \{n\} are \{positive/negative\}.' This mirrors the cognitive science finding that explicit denominators and comparison groups improve human statistical reasoning~\cite{gigerenzer1995improve}. \textbf{Numerical Anchoring:} By establishing specific frequency distributions before summarisation, REFER provides concrete numerical targets that guide attention allocation during generation, similar to how numerical frameworks reduce cognitive load in human judgement~\cite{tversky1974judgment}. \textbf{Sequential Deliberation:} REFER separates frequency analysis from summary generation, mirroring dual-process interventions that improve human reasoning by engaging deliberative rather than intuitive processing systems~\cite{evans2013dual}.
We propose five REFER strategies by incorporating existing state-of-the-art reasoning frameworks~\footnote{Note that the {direct prompt} template that we use in each of the following frameworks, outlined in Section~\ref{sec:compared_framework} on direct prompting, represents a typical baseline prompt used in opinion summarisation.}.

\begin{itemize}
\item\textbf{REFER} 
    we prompt the model to first analyse the frequency distribution of opinions across input documents, identifying how often specific opinions appear. By reasoning explicitly about these frequencies, the model establishes clear reference classes grounded in quantitative patterns rather than abstract probabilities. The model then generates a summary using the direct prompt, with opinions proportionally represented according to the identified frequency distribution. This frequency-framed approach aligns with humans' evolutionarily developed cognitive mechanisms for processing quantitative information, helping ensure fair and balanced generation.
        \begin{bluebox}[]{REFER}
        Let's first determine how many reviews out of \{number of input in the source documents\} are \{social value 1, social value 2\}. Then, let's generate a balanced summary that accurately reflects the frequency of these opinions. \{direct prompt\}
        \end{bluebox}

\item\textbf{Prefix-Instruct REFER (Prefix-Instruct-R)} we employ instructions by directly adding a prefix to instruct the model to be fair~\cite{borchers2022looking}. The aim of this explicit fairness instruction is to improve the model's fairness in its outputs through direct instruction. Additionally, we incorporate REFER by instructing frequency prompt in the direct instruction. The overall prompt thus becomes `be fair with your output \{REFER\}'.

\item\textbf{Prefix Role REFER (Prefix-Role-R)} numerous studies have revealed the benefits of introducing personas in language models~\cite{pataranutaporn2021ai, wang2023rolellm}. 
We instruct language models to act as fair individuals and incorporate REFER by instructing frequency prompt in the persona instruction. The overall prompt thus becomes `You are an unbiased person. \{REFER\}'.

\item\textbf{Chain of Thought REFER (CoT REFER)} Chain of Thought (CoT)~\cite{wei2022chain} is a reasoning technique that helps language models produce more accurate answers by breaking down complex problems into step-by-step thinking. We instruct language models to use CoT and incorporate REFER by instructing frequency prompt in the reasoning instruction. The overall prompt thus becomes `Let's think step by step. \{REFER\}'.

\item\textbf{Agent collaboration REFER (Agent-R)} research has demonstrated how LLMs can function as collaborative agents in multi-agent systems, enabling distributed problem-solving~\cite{liang2023encouraging, long2024multi, zhang2024chain}. We extend this approach in our summarisation workflow by assigning agents to different professional roles. These agents work together iteratively, each building upon the previous agent's output to refine the final summary.
        
    \textbf{Summarisation agent-REFER}  specialises in condensing opinions into concise and accurate summaries. By incorporating REFER, it analyses diverse viewpoints across multiple reviews, identifying key sentiment patterns and extracting the most representative perspectives and then generates summaries based on the frequency information.
    \textbf{Frequency agent} specialises in analysing and classifying opinions into appropriate categories to compute opinion frequencies.
    \textbf{Judge agent} evaluates the alignment between the summary generated by the summarisation agent and the frequency distribution provided by the frequency agent. Its task is to verify the summary's accuracy and provide constructive feedback for future improvements.
    \textbf{Senior editor agent} reviews and revises summaries based on the judge's feedback to ensure they accurately reflect the underlying opinion frequency distribution. This process serves as the final quality control mechanism for maintaining fair representation of opinions in the output summaries.
\end{itemize}
The full prompt can be found in Appendix~\ref{sec:agent_refer_prompt}.

\section{Experiments}
\subsection{Datasets}
\label{sec:datasets}
In our experimental setup, we use two datasets: FairSumm~\cite{dash2019summarizing} and Amazon Reviews 2023~\cite{hou2024bridging}.~\footnote{\url{https://huggingface.co/datasets/McAuley-Lab/Amazon-Reviews-2023}} To evaluate model fairness, we manually sampled and curated test sets for both political tweet and review summarisation tasks from existing datasets. For the Amazon Reviews dataset, we select reviews for the same product, containing between 30 and 120 words. The test sets are structured to contain 900 input collections each, with individual collections comprising either 30 political tweets or 8 reviews. To assess model fairness across varying input distributions, we implement three distinct input proportions: balanced input (i.e. equal representation from both values), and two skewed input scenarios (asymmetric distribution with 75\% social value 1 and 25\% social value 2, and vice versa). Each input proportion category contains 300 collections, resulting in a total of 900 input collections for comprehensive evaluation.

\subsection{Compared Frameworks}
\label{sec:compared_framework}

Various prompting frameworks have been studied to enhance language models' reasoning capabilities. We compare our proposed REFER frameworks with the following frameworks, which serve as baseline frameworks for our evaluation.

\noindent\textbf{Direct prompting} the most basic prompt we use to directly instruct the model to generate a summary from the input documents. The output serves as our overall fairness baseline by using the prompt: "Reviews about {topic}. Each review is separated by ||: {source}. Please write a short text containing the salient information (i.e., a summary). The summary of the reviews is:".
    
\noindent\textbf{Fair prefix prompting} following~\cite{borchers2022looking} we add instructions by directly adding the prefix "be fair with your output" to the direct prompt. The goal of this explicit fairness instruction is to improve the model's fairness in its outputs through direct instruction.
    
\noindent\textbf{Persona prefix prompting} numerous studies have revealed the benefits of introducing personas in language models~\cite{pataranutaporn2021ai, wang2023rolellm}. 
    In this approach, we instruct language models to act as fair individuals by prepending "You are an unbiased person" to the direct prompt, which has been shown to improve fairness in the generated output.

\noindent\textbf{Zero-shot CoT}~\cite{wei2022chain} we prepend basic CoT instructions "Let's think step by step" to the direct prompt. This guides the model in generating summaries step by step and breaks down the summarisation task into smaller, more manageable steps.

\noindent\textbf{Agent collaboration (Agent)} a based version of Agent collaboration REFER is implemented by using a simple prompt for the summarisation agent: "You are a summarisation specialist with exceptional ability to distil complex information into clear, concise, and accurate key points while preserving essential context and meaning.".
     
    



\subsection{Models}
We experiment with several state-of-the-art LLMs, including both open-source and proprietary models. For open-source models, we use Llama 3, specifically the instruct-tuned version 3.1 in both 8B~\footnote{\url{https://huggingface.co/meta-llama/Llama-3.1-8B-Instruct}} and 70B~\footnote{\url{https://huggingface.co/meta-llama/Llama-3.1-70B-Instruct}} parameter sizes. For proprietary models, we use three popular GPT models: GPT-3.5-Turbo (gpt-3.5-turbo-0125), GPT-4o-mini (gpt-4o-mini-2024-07-18), and GPT-4o (gpt-4o-2024-08-06). The implementation details can be found in Appendix~\ref{sec:implementation_details}.

\subsection{Evaluation Metrics}
\begin{table*}[htbp]
\begin{subtable}{\textwidth}
    \centering
    \setlength{\tabcolsep}{1.4pt}  
    \tiny
    \begin{tabular}{l c c c c c c c c c c c c c c c c c c c c}
        \toprule
        Model & \multicolumn{4}{c}{GPT-3.5-Turbo} & \multicolumn{4}{c}{GPT-4o-mini} & \multicolumn{4}{c}{GPT-4o} & \multicolumn{4}{c}{Llama 3-8B} & \multicolumn{4}{c}{Llama 3-70B} \\
        \cmidrule(lr){2-5} \cmidrule(lr){6-9} \cmidrule(lr){10-13} \cmidrule(lr){14-17} \cmidrule(lr){18-21}
        Metrics & \multicolumn{1}{c}{SPD$\downarrow$} & \multicolumn{1}{c}{BUR$\downarrow$} & \multicolumn{1}{c}{UER$\downarrow$} & \multicolumn{1}{c}{SOF$\downarrow$} & \multicolumn{1}{c}{SPD$\downarrow$} & \multicolumn{1}{c}{BUR$\downarrow$} & \multicolumn{1}{c}{UER$\downarrow$} & \multicolumn{1}{c}{SOF$\downarrow$} & \multicolumn{1}{c}{SPD$\downarrow$} & \multicolumn{1}{c}{BUR$\downarrow$} & \multicolumn{1}{c}{UER$\downarrow$} & \multicolumn{1}{c}{SOF$\downarrow$} & \multicolumn{1}{c}{SPD$\downarrow$} & \multicolumn{1}{c}{BUR$\downarrow$} & \multicolumn{1}{c}{UER$\downarrow$} & \multicolumn{1}{c}{SOF$\downarrow$} & \multicolumn{1}{c}{SPD$\downarrow$} & \multicolumn{1}{c}{BUR$\downarrow$} & \multicolumn{1}{c}{UER$\downarrow$} & \multicolumn{1}{c}{SOF$\downarrow$}  \\
        \midrule
        Direct Prompting & 35.07 & 57.59 & 8.07 & 6.66 & 30.77 & 54.67 & 7.84 & 6.53 & 28.43 & 55.37 & 7.86 & 6.52 & 27.97 & 57.19 & 7.94 & 6.22 & 26.87 & 58.22 & 8.09 & 6.47 \\
        REFER & \textcolor{blue}{(-4.07)} & \textcolor{orange}{(+1.22)} & \textcolor{orange}{(+0.17)} & \textcolor{blue}{(-0.10)} & \textcolor{blue}{(-3.50)} & \textcolor{blue}{(-1.37)} & \textcolor{orange}{(+0.05)} & \textcolor{blue}{(-0.02)} & \textcolor{orange}{(+1.04)} & \textcolor{orange}{(+1.33)} & \textcolor{orange}{(+0.10)} & (+0.00) & \textcolor{orange}{(+5.90)} & \textcolor{orange}{(+5.03)} & \textcolor{orange}{(+0.66)} & \textcolor{orange}{(+0.45)} & \textcolor{orange}{(+0.50)} & \textcolor{orange}{(+1.45)} & \textcolor{orange}{(+0.07)} & \textcolor{blue}{(-0.05)} \\
        \midrule
        Prefix-instruction & 34.80 & 56.56 & 8.02 & 6.66 & 30.67 & 54.67 & 7.82 & 6.52 & 29.70 & 55.22 & 7.87 & 6.60 & 28.60 & 59.33 & 8.16 & 6.44 & 27.57 & 58.19 & 8.06 & 6.47 \\
        Prefix-instruct-R & \textcolor{blue}{(-2.63)} & \textcolor{blue}{(-0.60)} & \textcolor{blue}{(-0.19)} & \textcolor{blue}{(-0.37)} & \textcolor{blue}{(-3.20)} & \textcolor{blue}{(-1.37)} & \textcolor{blue}{(-0.21)} & \textcolor{blue}{(-0.19)} & \textcolor{blue}{(-1.17)} & \textcolor{blue}{(-1.11)} & \textcolor{blue}{(-0.16)} & \textcolor{blue}{(-0.25)} & \textcolor{orange}{(+3.13)} & \textcolor{blue}{(-1.37)} & \textcolor{blue}{(-0.14)} & \textcolor{blue}{(-0.01)} & \textcolor{blue}{(-0.34)} & \textcolor{blue}{(-1.49)} & \textcolor{blue}{(-0.19)} & \textcolor{blue}{(-0.20)} \\
        \midrule
        Prefix-role & 34.73 & 57.33 & 8.00 & 6.62 & 30.53 & 55.22 & 7.82 & 6.54 & 28.50 & 54.93 & 7.90 & 6.59 & 26.80 & 57.04 & 7.90 & 6.16 & 28.53 & 59.00 & 8.13 & 6.44 \\
        Prefix-role-R & \textcolor{blue}{(-3.73)} & \textcolor{blue}{(-1.00)} & \textcolor{blue}{(-0.12)} & \textcolor{blue}{(-0.28)} & \textcolor{blue}{(-3.20)} & \textcolor{blue}{(-1.89)} & \textcolor{blue}{(-0.19)} & \textcolor{blue}{(-0.20)} & \textcolor{blue}{(-0.17)} & \textcolor{blue}{(-0.60)} & \textcolor{blue}{(-0.14)} & \textcolor{blue}{(-0.19)} & \textcolor{orange}{(+4.07)} & \textcolor{blue}{(-0.08)} & \textcolor{orange}{(+0.02)} & \textcolor{orange}{(+0.20)} & \textcolor{orange}{(+0.67)} & \textcolor{blue}{(-3.67)} & \textcolor{blue}{(-0.32)} & \textcolor{blue}{(-0.09)} \\
        \midrule
        CoT & 34.80 & 57.44 & 7.99 & 6.63 & 30.17 & 55.15 & 7.82 & 6.52 & 28.17 & 54.89 & 7.85 & 6.56 & 25.53 & 56.15 & 7.76 & 6.12 & 26.97 & 58.15 & 8.03 & 6.46 \\
        CoT-R & \textcolor{blue}{(-4.07)} & \textcolor{blue}{(-0.77)} & \textcolor{blue}{(-0.14)} & \textcolor{blue}{(-0.34)} & \textcolor{blue}{(-4.30)} & \textcolor{blue}{(-3.71)} & \textcolor{blue}{(-0.38)} & \textcolor{blue}{(-0.33)} & \textcolor{blue}{(-0.17)} & \textcolor{blue}{(-2.15)} & \textcolor{blue}{(-0.21)} & \textcolor{blue}{(-0.23)} & \textcolor{orange}{(+4.44)} & \textcolor{orange}{(+0.59)} & \textcolor{orange}{(+0.22)} & \textcolor{orange}{(+0.31)} & \textcolor{orange}{(+1.00)} & \textcolor{blue}{(-2.56)} & \textcolor{blue}{(-0.25)} & \textcolor{blue}{(-0.22)} \\
        \midrule
        Agent & 34.43 & 56.33 & 7.97 & 6.65 & 30.80 & 54.44 & 7.80 & 6.55 & 32.13 & 54.44 & 7.81 & 6.55 & 37.97 & 63.22 & 8.97 & 6.94 & 31.00 & 57.11 & 8.04 & 6.49 \\
        Agent-R & \textcolor{orange}{(+0.84)} & \textcolor{orange}{(+0.48)} & \textcolor{blue}{(-0.03)} & \textcolor{blue}{(-0.05)} & (+0.00) & \textcolor{orange}{(+1.56)} & \textcolor{orange}{(+0.07)} & \textcolor{blue}{(-0.03)} & \textcolor{blue}{(-3.36)} & \textcolor{orange}{(+0.12)} & (+0.00) & \textcolor{blue}{(-0.03)} & \textcolor{blue}{(-9.84)} & \textcolor{blue}{(-3.78)} & \textcolor{blue}{(-0.61)} & \textcolor{blue}{(-0.25)} & \textcolor{blue}{(-4.17)} & \textcolor{orange}{(+0.22)} & \textcolor{blue}{(-0.08)} & \textcolor{blue}{(-0.04)} \\
        \bottomrule
    \end{tabular}
    \caption{Review summarisation}
\end{subtable}

    \vspace{0.5em}  
\begin{subtable}{\textwidth}
    \centering
    \setlength{\tabcolsep}{1.4pt}  
    \tiny
    \begin{tabular}{l c c c c c c c c c c c c c c c c c c c c}
        \toprule
        Model & \multicolumn{4}{c}{GPT-3.5-Turbo} & \multicolumn{4}{c}{GPT-4o-mini} & \multicolumn{4}{c}{GPT-4o} & \multicolumn{4}{c}{Llama 3-8B} & \multicolumn{4}{c}{Llama 3-70B} \\
        \cmidrule(lr){2-5} \cmidrule(lr){6-9} \cmidrule(lr){10-13} \cmidrule(lr){14-17} \cmidrule(lr){18-21}
        Metrics & \multicolumn{1}{c}{SPD$\downarrow$} & \multicolumn{1}{c}{BUR$\downarrow$} & \multicolumn{1}{c}{UER$\downarrow$} & \multicolumn{1}{c}{SOF$\downarrow$} & \multicolumn{1}{c}{SPD$\downarrow$} & \multicolumn{1}{c}{BUR$\downarrow$} & \multicolumn{1}{c}{UER$\downarrow$} & \multicolumn{1}{c}{SOF$\downarrow$} & \multicolumn{1}{c}{SPD$\downarrow$} & \multicolumn{1}{c}{BUR$\downarrow$} & \multicolumn{1}{c}{UER$\downarrow$} & \multicolumn{1}{c}{SOF$\downarrow$} & \multicolumn{1}{c}{SPD$\downarrow$} & \multicolumn{1}{c}{BUR$\downarrow$} & \multicolumn{1}{c}{UER$\downarrow$} & \multicolumn{1}{c}{SOF$\downarrow$} & \multicolumn{1}{c}{SPD$\downarrow$} & \multicolumn{1}{c}{BUR$\downarrow$} & \multicolumn{1}{c}{UER$\downarrow$} & \multicolumn{1}{c}{SOF$\downarrow$}  \\
        \midrule
        Direct Prompting & 37.50 & 66.44 & 8.69 & 8.48 & 32.60 & 66.67 & 8.57 & 8.36 & 34.27 & 66.67 & 8.65 & 8.42 & 34.63 & 66.44 & 8.66 & 8.45 & 38.37 & 66.67 & 8.78 & 8.56 \\
        REFER & \textcolor{blue}{(-0.80)} & \textcolor{orange}{(+0.23)} & \textcolor{orange}{(+0.12)} & \textcolor{orange}{(+0.09)} & \textcolor{blue}{(-2.77)} & \textcolor{orange}{(+0.11)} & \textcolor{orange}{(+0.26)} & \textcolor{orange}{(+0.25)} & \textcolor{blue}{(-0.70)} & \textcolor{orange}{(+0.11)} & \textcolor{orange}{(+0.12)} & \textcolor{orange}{(+0.12)} & \textcolor{orange}{(+3.14)} & \textcolor{orange}{(+0.23)} & \textcolor{orange}{(+0.25)} & \textcolor{orange}{(+0.23)} & \textcolor{blue}{(-3.14)} & (+0.00) & \textcolor{blue}{(-0.01)} & \textcolor{blue}{(-0.02)} \\
        \midrule
        Prefix-instruct & 36.60 & 66.67 & 8.74 & 8.52 & 32.07 & 66.67 & 8.66 & 8.47 & 34.97 & 66.67 & 8.64 & 8.45 & 36.77 & 66.56 & 8.72 & 8.48 & 38.07 & 66.67 & 8.83 & 8.63 \\
        Prefix-instruct-R & \textcolor{orange}{(+1.37)} & (+0.00) & \textcolor{blue}{(-0.30)} & \textcolor{blue}{(-0.28)} & \textcolor{blue}{(-1.90)} & \textcolor{blue}{(-0.11)} & \textcolor{blue}{(-0.09)} & \textcolor{blue}{(-0.12)} & \textcolor{blue}{(-1.30)} & \textcolor{orange}{(+0.11)} & \textcolor{blue}{(-0.16)} & \textcolor{blue}{(-0.21)} & \textcolor{orange}{(+1.30)} & \textcolor{blue}{(-0.45)} & \textcolor{blue}{(-0.11)} & \textcolor{blue}{(-0.12)} & \textcolor{blue}{(-3.44)} & (+0.00) & \textcolor{blue}{(-0.31)} & \textcolor{blue}{(-0.33)} \\
        \midrule
        Prefix-role & 36.50 & 66.67 & 8.75 & 8.53 & 32.90 & 66.67 & 8.66 & 8.47 & 34.23 & 66.78 & 8.66 & 8.45 & 34.33 & 66.67 & 8.71 & 8.48 & 37.53 & 66.67 & 8.85 & 8.64 \\
        Prefix-role-R & \textcolor{orange}{(+1.47)} & (+0.00) & \textcolor{blue}{(-0.31)} & \textcolor{blue}{(-0.31)} & \textcolor{blue}{(-2.30)} & \textcolor{blue}{(-0.56)} & \textcolor{blue}{(-0.20)} & \textcolor{blue}{(-0.19)} & \textcolor{blue}{(-0.30)} & \textcolor{blue}{(-0.56)} & \textcolor{blue}{(-0.24)} & \textcolor{blue}{(-0.27)} & \textcolor{orange}{(+3.97)} & \textcolor{blue}{(-0.34)} & \textcolor{blue}{(-0.08)} & \textcolor{blue}{(-0.08)} & \textcolor{blue}{(-3.66)} & (+0.00) & \textcolor{blue}{(-0.26)} & \textcolor{blue}{(-0.28)} \\
        \midrule
        CoT & 36.20 & 66.44 & 8.68 & 8.46 & 32.57 & 66.67 & 8.58 & 8.38 & 35.00 & 66.67 & 8.60 & 8.40 & 37.17 & 66.67 & 8.67 & 8.47 & 38.37 & 66.67 & 8.87 & 8.67 \\
        CoT-R & \textcolor{orange}{(+0.50)} & \textcolor{orange}{(+0.23)} & \textcolor{blue}{(-0.21)} & \textcolor{blue}{(-0.21)} & \textcolor{blue}{(-2.47)} & \textcolor{blue}{(-1.11)} & \textcolor{blue}{(-0.14)} & \textcolor{blue}{(-0.26)} & \textcolor{blue}{(-0.83)} & \textcolor{orange}{(+0.44)} & \textcolor{blue}{(-0.11)} & \textcolor{blue}{(-0.16)} & \textcolor{orange}{(+0.66)} & \textcolor{blue}{(-2.11)} & \textcolor{blue}{(-0.27)} & \textcolor{blue}{(-0.31)} & \textcolor{blue}{(-5.34)} & \textcolor{orange}{(+0.11)} & \textcolor{blue}{(-0.51)} & \textcolor{blue}{(-0.54)} \\
        \midrule
        Agent & 36.40 & 66.67 & 8.68 & 8.44 & 31.87 & 66.67 & 8.56 & 8.37 & 31.40 & 66.33 & 8.44 & 8.22 & 36.30 & 66.67 & 8.83 & 8.59 & 32.23 & 66.67 & 8.75 & 8.57 \\
        Agent-R & \textcolor{blue}{(-5.27)} & (+0.00) & \textcolor{orange}{(+0.01)} & \textcolor{orange}{(+0.03)} & \textcolor{blue}{(-1.44)} & (+0.00) & \textcolor{blue}{(-0.05)} & \textcolor{blue}{(-0.06)} & \textcolor{orange}{(+0.87)} & \textcolor{orange}{(+0.23)} & \textcolor{blue}{(-0.02)} & (+0.00) & (+0.00) & \textcolor{blue}{(-0.34)} & \textcolor{blue}{(-0.09)} & \textcolor{blue}{(-0.12)} & \textcolor{blue}{(-1.26)} & (+0.00) & \textcolor{orange}{(+0.03)} & \textcolor{orange}{(+0.01)} \\
        \bottomrule
    \end{tabular}
    \caption{Political tweet summarisation}
\end{subtable}
    \caption{
    Fairness evaluation comparing different incorporations of REFER into existing frameworks. Original values are shown with REFER differences in brackets on the second line. Lower values ($\downarrow$) indicate better fairness. \textcolor{blue}{Blue} and (+) shows improvements, \textcolor{orange}{Orange} and (-) shows regressions.
    }  
\label{tab:frequency_prompt_fairness}
\end{table*}

We evaluate model fairness by comparing opinion distributions in generated summaries against source documents, focusing on proportional representation~\cite{shandilya2018fairness}. 
We use four different metrics: Second-Order SPD (SPD)~\cite{huang-etal-2024-bias}, Binary Unfair Rate (BUR), Unfair Error Rate (UER), and Second-Order Fairness (SOF)~\cite{zhang2023fair}.
SPD evaluates fairness by classifying social attributes at the sentence level in summaries, then comparing these distributions against source documents. The other metrics compare value distributions through token-based approaches: BUR quantifies the ratio of fair summaries to total generated summaries, UER measures underrepresentation by calculating discrepancies between target and generated social value distributions, and SOF assesses the variance of unfairness across different social values within each sample.

Since model-generated summaries often contain compound sentences with multiple opinions, we first use GPT-4o-mini to decompose them into single-opinion statements by prompting "Split the following sentences into simple propositions without introducing new information, do it sentence by sentence: \textbackslash n\textbackslash n Sentences: \{model generated summary\}". We then apply the evaluation metrics to these sentences that carry single opinion.
Following~\citet{huang-etal-2024-bias}, we calculate SPD using their classification approach. For BUR, UER, and SOF, we use the BARTScore~\cite{yuan2021bartscore} implementation proposed by~\citet{zhang2023fair}, as it demonstrates stronger alignment with human judgement.

\section{Results and Discussion}

\subsection{Overall Impact of REFER on Fairness}
\label{sec:basic_fairness_result}
Table~\ref{tab:frequency_prompt_fairness} presents a comprehensive evaluation of fairness across different prompting strategies incorporated with REFER, assessed on two summarisation datasets: political tweet summarisation and review summarisation. The evaluation considers four fairness metrics and their absolute values, including SPD, BUR, UER, and SOF, where lower values indicate improved fairness. Results are averaged across different sets of input distributions mentioned in Section~\ref{sec:datasets}.

First-order fairness metrics, including BUR and UER, reflect the equitable distribution of model outputs across social attributes. Most frameworks and their REFER counterparts have relatively similar BUR values, especially when summarising political tweets. For UER, the majority of models have REFER counterparts that outperform or show comparable values, except for Llama 3-8B. This suggests that while most frameworks achieve similar fairness in broad representation (BUR), REFER generally helps improve representation across attributes (UER), indicating its effectiveness at enhancing fine-grained fairness in opinion summarisation.



Second-order metrics (SPD and SOF) detect subtle, systematic biases by accounting for input social value distributions. A good second-order fairness means that the summary maintains similar patterns of differences between groups as the source text. These metrics help us understand whether a model is consistently biased against particular groups or randomly unfair, a distinction crucial for improving models. Notably, more models and their REFER counterparts achieve better second-order fairness, especially on the review dataset. Across different models and input datasets, REFER variants incorporating structured reasoning prompts (Prefix-instruct-R, Prefix-role-R, and CoT-R) show improvements in second-order fairness metrics across most model-dataset combinations, though the magnitude of improvement varies considerably by model architecture and size.


Table~\ref{tab:frequency_prompt_fairness} shows that REFER exhibits reduced effectiveness when deployed with smaller models such as Llama 3-8B, while demonstrating strong performance when applied to larger language models. 
This can be due to larger models developing improved internal mechanisms for language processing that make them more efficient in representing and generating information~\cite{zhao2024explainability, lindsey2025biology}.
In contrast, smaller models are not as effective at following complex instructions~\cite{qin2024infobench, ouyang2022training}. Smaller models such as Llama 3-8B appear to exhibit more variability in output patterns and tend to generate summaries with varying lengths, as evidenced in the summary length analysis in Appendix~\ref{sec:summary_length}. This variability introduces more randomness in the summaries, making them harder to compare.

Our qualitative analysis in Appendix~\ref{sec:summary_length} demonstrates this through direct comparison of CoT-REFER outputs between Llama 3-8B and Llama 3-70B variants. When instructed to first calculate frequency information before summarising product reviews, the 8B model completely ignored the sequential instruction requirement and immediately proceeded with qualitative assessment, while the 70B model properly executed the instruction by beginning with precise quantification before providing summary analysis. This suggests that REFER's effectiveness can be inherently linked to a model's fundamental ability to process and respond to complex instructions~\cite{kim2024biggen}.

\begin{figure*}[htbp]
    \centering
    
    \begin{subfigure}[t]{0.93\textwidth}
        \centering
        \includegraphics[width=\linewidth]{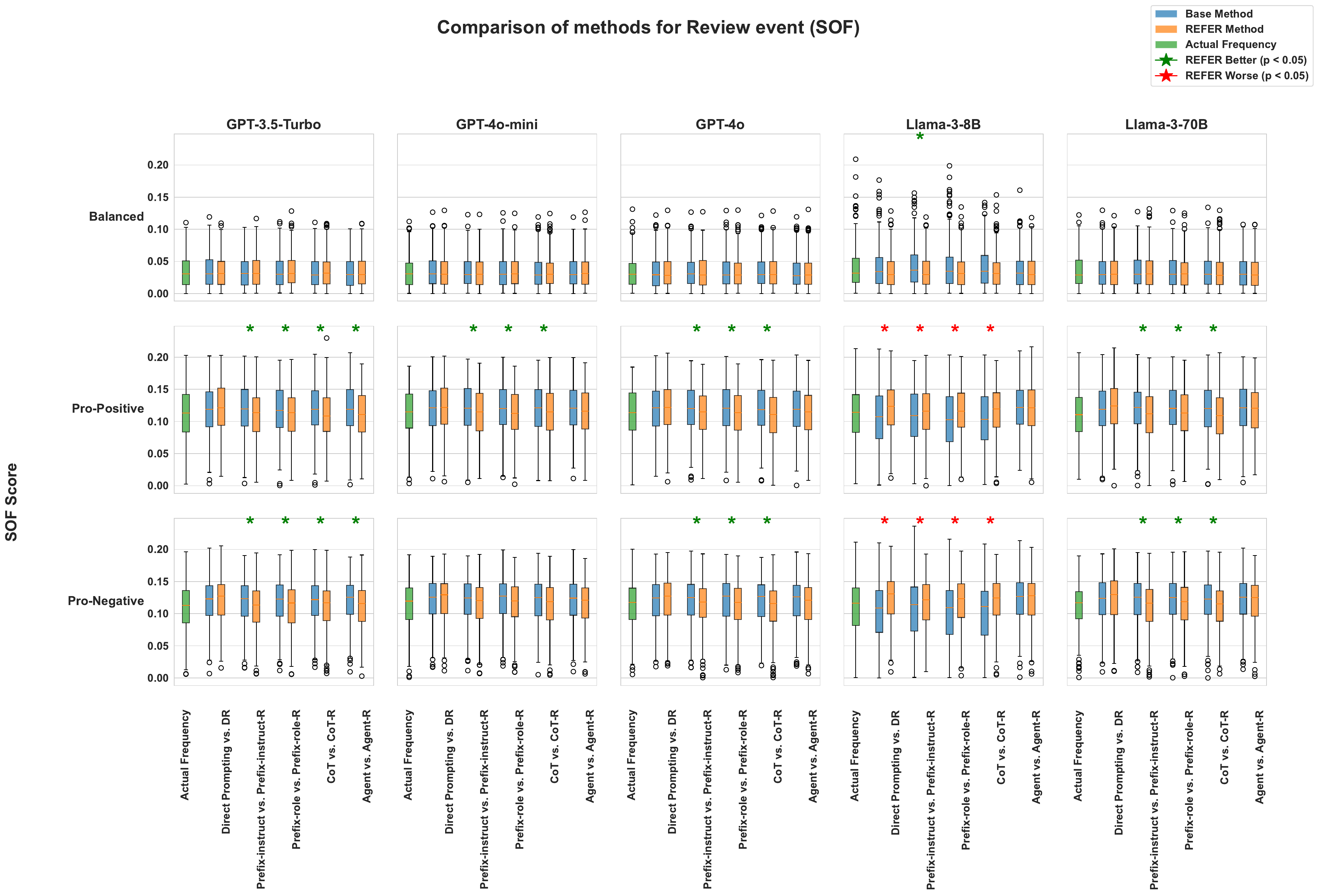}
        \caption{Review summarisation - SOF}
        \label{fig:sof_review}
    \end{subfigure}
    
    \vspace{0.1cm} 
    
    \begin{subfigure}[t]{\textwidth}
        \centering
        \includegraphics[width=0.93\linewidth]{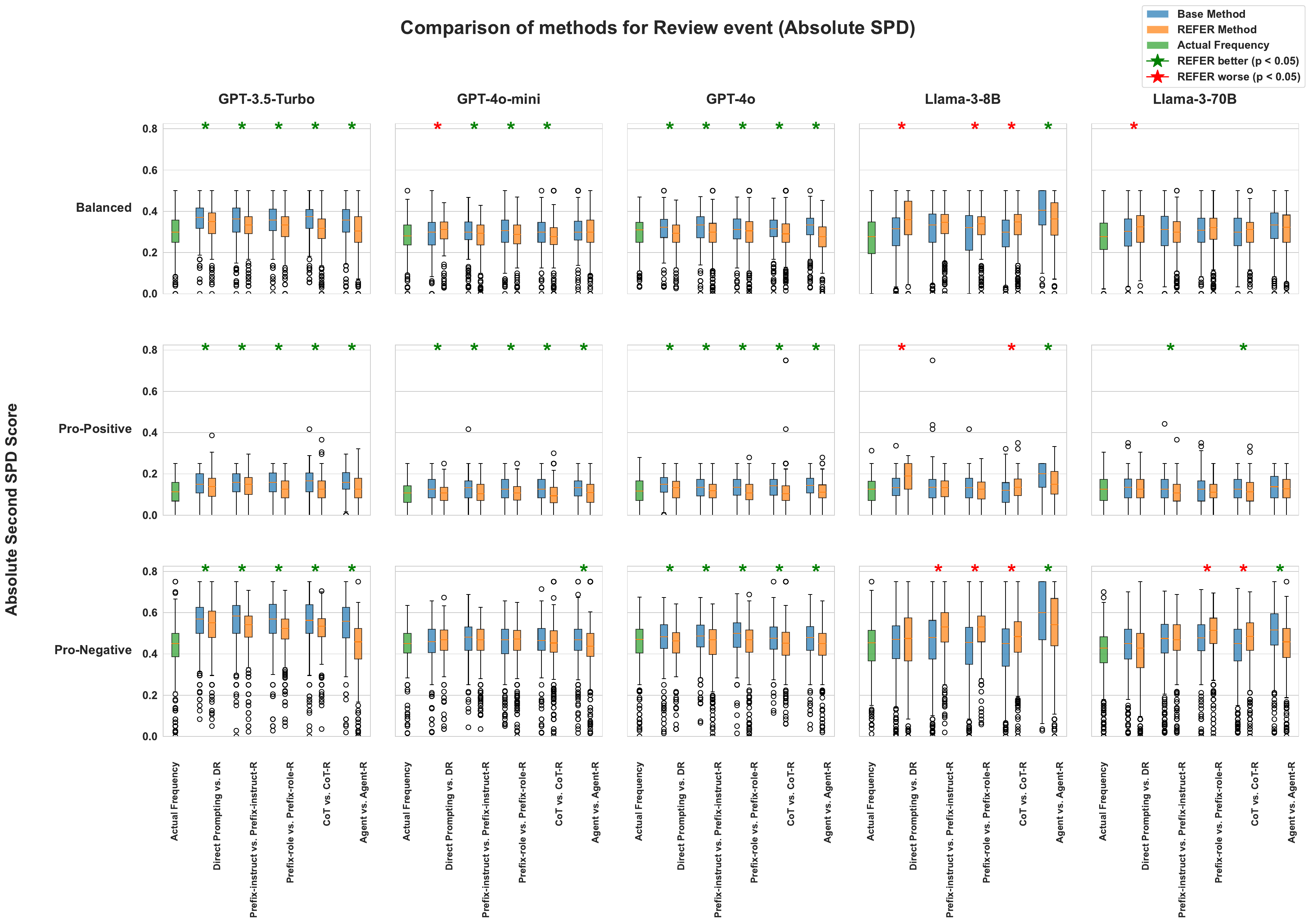}
        \caption{Review summarisation - SPD}
        \label{fig:spd_review}
    \end{subfigure}
    
    \caption{Comparison of SOF and SPD metrics for review summarisation. Green bars represent the oracle prompt by including actual frequency information. Blue bars are the base frameworks and orange bars are the REFER counterparts associated with them. When a REFER framework is statistically significantly better (lower value) than its base framework, the pair is highlighted using a green star on top. If a base framework is better, then it is highlighted using a red star.}
    \label{fig:review_metrics}
\end{figure*}

Overall, models are less biased according to all metrics when summarising reviews compared to political tweets. REFER-enhanced prompting improves fairness, with Prefix-instruct-R, Prefix-role-R and CoT-R being particularly effective, exhibiting balanced performance across both first-order and second-order fairness metrics. These findings suggest REFER's benefit across model reasoning processes when generating summaries, highlighting potential to improve LLM-generated content across varied architectures and datasets. Due to performance issues identified in the analysis above, we exclude Llama 3-8B from further analysis.


\subsection{Second-Order Fairness Patterns}

In this section, we examine second-order fairness in opinion summarisation in greater detail by examining different input proportions rather than using aggregated results, and conducting statistical analyses. While first-order metrics such as BUR and UER measure whether all groups are represented, second-order metrics such as SPD and SOF examine whether the proportional relationships between groups are preserved. This distinction is crucial because a model that consistently underrepresents minority opinions by the same margin across different contexts exhibits systematic bias, which is more problematic than random variations in representation.


Using datasets and input proportions from Section~\ref{sec:datasets}, we also include an oracle result with actual frequency prompts: `\{number 1\} and \{number 2\} out of \{number of input\} are \{social value 1, social value 2\}, generate a balanced summary reflecting this distribution. \{direct prompt\}'. The oracle prompt, which provides exact frequency information, establishes a theoretical upper bound for fairness performance. 
We report results using base frameworks and their REFER counterparts, with Mann-Whitney U tests for statistical significance. In visualisations, significantly better REFER results are highlighted with green stars. Review dataset results appear in Figure~\ref{fig:review_metrics}, showing patterns similar to the political tweets dataset, with full results in Appendix~\ref{sec:second_order_fairness_full_results}.

In our analysis, the prompt incorporating actual frequencies proved most fair according to both metrics---an unsurprising result that serves as an upper bound for fairness achievement. 
REFER demonstrated statistically significant effectiveness when combined with Chain-of-Thought reasoning, prefix-based instructions, and persona-driven prompting, yielding the best overall performance across most LLMs and proving particularly effective with skewed input distributions. Chain-of-Thought REFER explicitly forces models to work through frequency calculations before summarisation, creating deliberative pathways that mirror human statistical reasoning. Prefix-instruction and Prefix-role REFER leverage instruction-following capabilities while providing numerical anchors for fairness assessment. 
Conversely, Agent collaboration REFER showed mixed results, suggesting that REFER is most effective when executed through unified reasoning processes rather than the added complexity of multi-party collaboration. This finding aligns with the original theoretical development of frequency-based reasoning frameworks, which were grounded in individual cognitive processes rather than distributed reasoning paradigms~\citep{tversky1983extensional, fiedler1988dependence, gigerenzer1994distinction, gigerenzer1995improve, gigerenzer1999overcoming}.

Notably, transitioning to REFER produced larger shifts in fairness metrics than choosing between different base prompting frameworks. As visualised in Figure~\ref{fig:review_metrics}, the variance amongst base frameworks (blue bars) is considerably smaller than the improvements achieved through REFER implementation (orange bars). This suggests that frequency-based reasoning may be more fundamental to fairness than the specific prompting strategy employed.

\section{Conclusion}
This study introduces REFER (Frequency Framed Prompting), a cognitive science-inspired debiasing framework for opinion summarisation. Drawing from research showing humans make more accurate statistical judgements with concrete frequencies rather than abstract probabilities, we demonstrate that frequency-based prompting enhances fairness in LLM-generated summaries.
Our evaluation reveals that REFER improves model fairness across multiple datasets, models, and metrics, particularly when combined with stronger reasoning instructions and in larger models with superior instruction-following capabilities. Notably, REFER improvements often exceed differences between base prompting frameworks, suggesting frequency-based reasoning plays a greater role than specific prompting strategies in addressing fairness.
REFER offers a lightweight, end-user accessible solution that works with both open-source and proprietary models without requiring retraining or hyperparameter tuning, addressing critical gaps in existing computationally demanding debiasing approaches.


\section*{Limitations}
This study specifically focuses on model fairness rather than addressing dataset bias. Given that the fairness characteristics of the models' training data are unknown, our analysis is confined to examining the fairness of the generated summarisation outputs. The prompts employed in this study were manually crafted without extensive optimisation procedures. While our primary focus was on evaluating the discrete steps, eliciting frequency-based responses and analysing components of the summarisation process rather than prompt optimisation, we acknowledge that determining the optimal prompt formulation remains an ongoing challenge. Given computational and resource constraints, we conducted selective testing of representative prompting frameworks rather than an exhaustive evaluation of all possible approaches. Nevertheless, our approach presents a generalisable framework for utilising LLMs to achieve efficient debiasing. Future research directions could incorporate automated prompt generation techniques to further enhance the effectiveness of our methodology.

This study draws parallels between human cognitive processes and LLM information processing based on observed behavioural similarities. However, we acknowledge that these comparisons are fundamentally limited by our incomplete understanding of both human cognition and LLM internal mechanisms. The frequency-based reasoning improvements we observe in LLMs may arise from entirely different computational processes than those underlying human statistical reasoning, despite producing similar outcomes. Our analogy to human cognitive science serves as a useful framework for developing prompting strategies rather than making claims about mechanistic equivalence between human and artificial intelligence.

\section*{Ethical Considerations}
This study followed ethical principles and guidelines. 
The authors of this paper by no means suggest that language models are intentionally biased. 
This study builds on extensive literature demonstrating that LLMs exhibit various forms of bias due to patterns in their training data. While these biases are not intentionally designed into the systems, they nonetheless have real impacts on users and society. Our work aims to provide practical methods for mitigating these well-documented biases in the specific context of opinion summarisation.
We highly encourage readers to investigate and evaluate the findings for themselves. 
Overall, the goal of our research is to promote awareness of bias in summarising social media text since it is critical to understand what is summarised and whether it represents actual public opinions. Our work contributes to understanding the biases of summarisation models when summarising social media text, which is crucial for ethical use.

Our approach relies on predefined labels in datasets to measure bias. These labels are assigned based on established policies. However, if the labelling policy itself is inaccurate, our procedure might measure bias incorrectly. Therefore, we recommend using our technique only with datasets that have undergone careful review and construction to ensure accurate labelling.

\bibliography{custom}
\appendix

\section{Appendix}
\label{sec:appendix}
\subsection{Agent REFER Prompt}
\label{sec:agent_refer_prompt}

        \begin{bluebox}[]{Summarisation Agent-REFER}
        You are a summarisation specialist with exceptional ability to distil complex information into clear, concise, and accurate key points whilst preserving essential context and meaning. Let's first determine how many reviews out of {number of input in the source documents} are {social value 1, social value 2}. Then, let's generate a balanced summary that accurately reflects this distribution of opinions. {source input}
        
        \vspace{8pt}
        
        First, provide the counts in this format: {social value 1 \#number, social value 2 \#number}
        
        \vspace{8pt}
        
        Please write a short text containing the salient information that reflects this distribution. The summary of the reviews is:
        \end{bluebox}

        \begin{bluebox}[]{Frequency Agent}
        You are a classification expert with superb ability to analyse, categorise, and sort any type of information with exceptional precision and accuracy. Classify how many reviews out of \{number of input in the source documents\} are \{social value 1, social value 2\}. Reviews to analyse: \{source input\} 
                
        \vspace{8pt}

        Output exactly in format: \{social value 1 \#number, social value 2 \#number\}
        \end{bluebox}
        
        \begin{bluebox}[]{Judge Agent}
        You are a validation expert specialising in quality control of summaries. You compare summaries against source opinion frequency distributions, precisely identify discrepancies, and provide targeted recommendations for alignment and improvement. \{source input\} 
        
        \vspace{8pt}

        Opinion frequency distribution: \{frequency classification by the Frequency Agent\}
        
        \vspace{8pt}

        Summary: \{summary generated by the Summarisation Agent\} 
        
        \vspace{8pt}

        Compare the summary against the opinion frequency distribution, checking for bias and misrepresentation.
        \end{bluebox}
        
        \begin{bluebox}[]{Senior Editor Agent}
        You are a senior editor who expertly revises summaries based on detailed validation feedback, ensuring perfect alignment with source opinion frequency distributions while maintaining clarity and conciseness. You operate as the final quality control gate, transforming flagged summaries into highly accurate versions. \{source input\} 
        
        \vspace{8pt}

        Opinion frequency distribution:  \{frequency classification by the Frequency Agent\} 
        
        \vspace{8pt}
  
        Summary: \{summary generated by the Summarisation Agent\}
        
        \vspace{8pt}

        Validation feedback: \{validation and feedback generated by the Judge Agent\}
        
        \vspace{8pt}
        
        Revise the summary to align with the opinion frequency distribution while maintaining clarity and balance.
        \end{bluebox}
        
\subsection{Implementation Details}
\label{sec:implementation_details}
We adopt zero-shot prompting for our experiments, as previous studies demonstrated that decoder-only instruction-tuned models perform effectively as zero-shot abstractive summarisers \cite{tang2023context, laskar2023building, adams2023sparse}.  
For open-source models, we utilised the model implementations and weights available from Hugging Face~\cite{wolf-etal-2020-transformers}, while proprietary models were accessed through their respective APIs. The experiments with open-source models were conducted using four NVIDIA A100 (40GB) GPUs. The hyperparameters for models employ strictly controlled settings during inference, including constrained output length with maximum new tokens of 256, low temperature of 0.001 for more reproducible output, and modest repetition mitigation via repetition penalty of 1.1, facilitating consistent and reproducible summarisation of review distributions.

\subsection{Summary Qualitative Analysis}
\label{sec:summary_length}
\begin{figure}[htbp]
    \centering
    \includegraphics[width=\columnwidth]{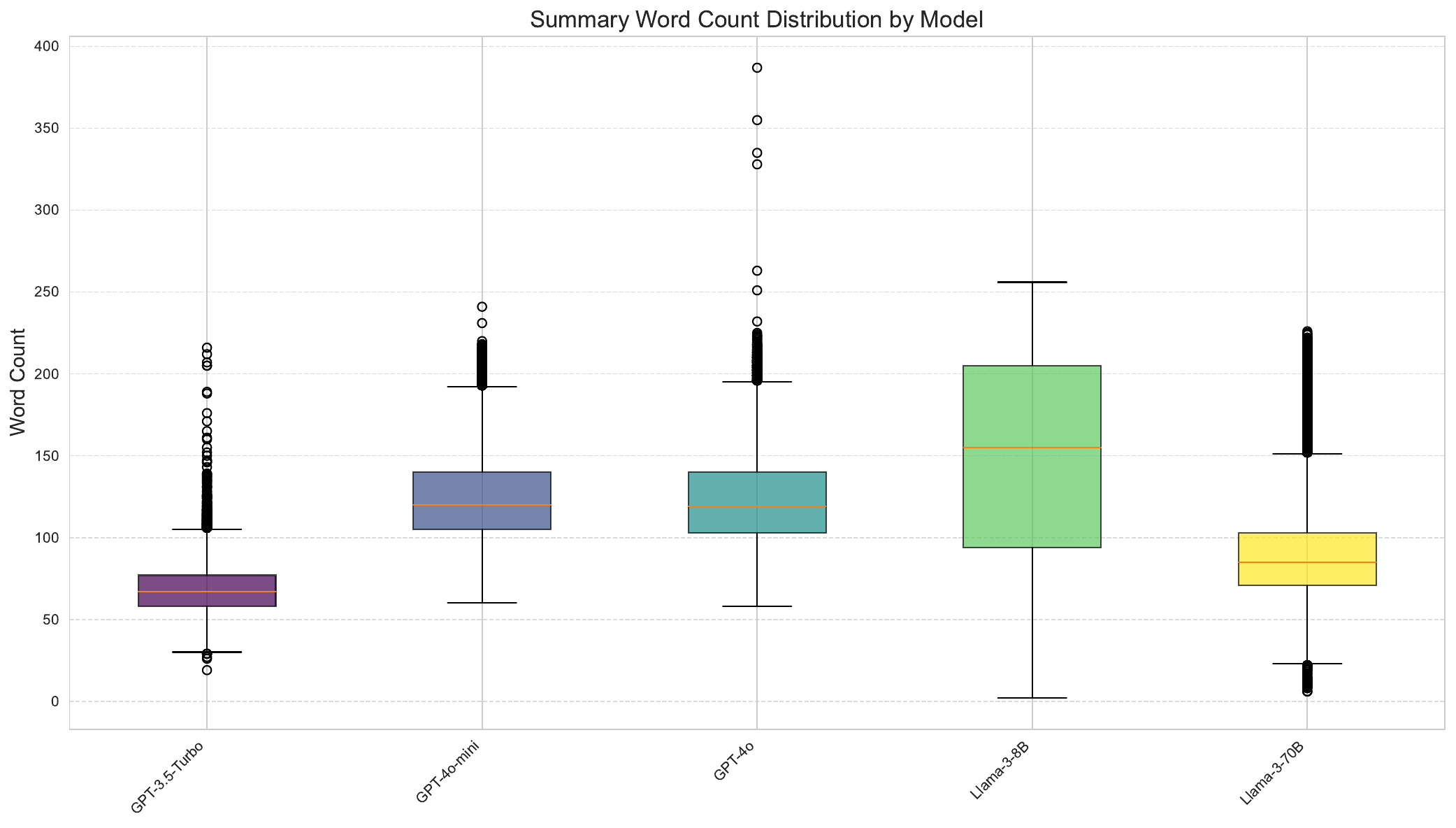}
    \caption{Summary word count distributions across different language models.}
    \label{fig:word_count_distributions}
\end{figure}
As visualised in Figure~\ref{fig:word_count_distributions}, the Llama 3-8B model exhibited the highest median word count among all models. Its interquartile range spanned approximately 95 to 205 words, indicating substantial variability in summary length within the central distribution. The extensive spread observed in the Llama3-8B distribution suggests a lack of consistency in generating summaries of uniform length, distinguishing it from other models which demonstrated comparatively more constrained distributions.

We also conducted a qualitative analysis across outputs generated using Llama3 and its 8B and 70B variants for instruction following capability. We present an example using the CoT-REFER prompt. Below are the original outputs from the two language models when instructed to first calculate frequency information before summarising product reviews:

\begin{tcolorbox}[title=Llama3-8B Output, colback=gray!5, colframe=gray!75!black]
``Overall, this product has mixed reviews. While some users found it effective, easy to install, and appreciated its affordable price and value, others reported problems with water flow and taste, and some returned it due to poor performance or installation difficulty. This indicates the product may work well for some but not for everyone, highlighting the importance of reading multiple reviews before deciding to purchase.''
\end{tcolorbox}

\begin{tcolorbox}[title=Llama3-70B Output, colback=gray!5, colframe=gray!75!black]
``Out of 8 reviews, 4 are positive and 4 are negative. Positive reviewers praise the product for being affordable, improving water taste and flow rate, and providing good value for the price. However, some reviewers experienced issues such as slow water flow, poor taste, and compatibility problems, leading them to return or replace the product. Overall, the product has mixed results, and potential buyers should be aware of the varying experiences reported by customers.''
\end{tcolorbox}

The analysis of these model outputs provides compelling evidence of how parameter scaling affects instruction-following capabilities aligns with previous studies~\cite{qin2024infobench, ouyang2022training}. The original instruction explicitly requested both models to first calculate frequency information before providing a summary of product reviews. As clearly demonstrated in the outputs and comparative table~\ref{tab:qualitative_analysis}, the 70B model successfully executed this sequential instruction, while the 8B parameter model did not. The 70B model begins with precise quantification---``Out of 8 reviews, 4 are positive and 4 are negative''---establishing the numerical distribution before proceeding to qualitative analysis. This suggests proper adherence to the instruction sequence. In contrast, the 8B model ignore the frequency calculation entirely, immediately beginning with a qualitative assessment: ``Overall, this product has mixed reviews.'' Both models capture similar content points regarding the product's performance, mentioning affordability, taste improvement, installation experience, and flow issues. However, the fundamental difference lies in their instruction-following capabilities rather than content comprehension. The larger model not only follows instructions more precisely but also presents information in a more structured manner with clear delineation between frequency analysis and subjective assessment.

\begin{table*}[htbp]
\centering
\renewcommand{\arraystretch}{1.3} 
\begin{tabular}{|p{4.5cm}|p{5cm}|p{5cm}|}
\hline
\textbf{Aspect} & \textbf{Llama3-70B} & \textbf{Llama3-8B} \\
\hline
Instruction Following & Explicitly follows the instruction to calculate frequency first: ``Out of 8 reviews, 4 are positive and 4 are negative.'' & Fails to follow the instruction sequence, omitting frequency analysis entirely. \\
\hline
Frequency Analysis & Quantifies the distribution precisely (4 positive, 4 negative) & No quantification provided \\
\hline
Summary Introduction & Begins with quantified data before transitioning to summary & Immediately begins with qualitative assessment: ``Overall, this product has mixed reviews.'' \\
\hline
Example of Positive Aspects & ``Positive reviewers praise the product for being affordable, improving water taste and flow rate, and providing good value for the price.'' & ``While some users found it effective, easy to install, and appreciated its affordable price and value...'' \\
\hline
Example of Negative Aspects & ``However, some reviewers experienced issues such as slow water flow, poor taste, and compatibility problems, leading them to return or replace the product.'' & ``...others reported problems with water flow and taste, and some returned it due to poor performance or installation difficulty.'' \\
\hline
\end{tabular}
\caption{Comparison of Instruction Following Between Llama3 8B and 70B Parameter Models}
\label{tab:qualitative_analysis}
\end{table*}

\subsection{Second Order Fairness Full Results}
\label{sec:second_order_fairness_full_results}
The raw results for both the political tweets summarisation and review summarisations are reported in Tables~\ref{tab:raw_sof_political}, \ref{tab:raw_sof_review}, \ref{tab:raw_spd_political}, and \ref{tab:raw_spd_review}.
REFER counterparts that perform better than the base framework are highlighted in green, while base framework results that are better are highlighted in red. The effects of REFER are more profound when input documents are skewed.

\begin{table*}[htbp]
\centering
\resizebox{\textwidth}{!}{
\begin{tabular}{llllll}
\toprule
Model & GPT-3.5-Turbo & GPT-4o-mini & GPT-4o & Llama-3-8B & Llama-3-70B \\
\multicolumn{6}{l}{\textbf{Balanced}}\\
Direct Prompting & 0.04 & 0.03 & 0.03 & 0.04 & 0.04 \\
REFER & 0.03 $\uparrow$(0.6\%) & 0.03 $\downarrow$(-0.4\%) & 0.03 $\downarrow$(-1.4\%) & 0.03 $\uparrow$(2.0\%) & 0.03 $\uparrow$(1.6\%) \\
Prefix-instruct & 0.03 & 0.03 & 0.03 & 0.03 & 0.04 \\
Prefix-instruct-R & 0.04 $\downarrow$(-3.7\%) & 0.03 $\uparrow$(0.9\%) & 0.03 $\uparrow$(3.0\%) & 0.03 $\uparrow$(1.3\%) & 0.04 $\downarrow$(-0.4\%) \\
Prefix-role & 0.03 & 0.03 & 0.03 & 0.04 & 0.04 \\
Prefix-role-R & 0.04 $\downarrow$(-2.0\%) & 0.03 $\downarrow$(-0.6\%) & 0.03 $\uparrow$(0.9\%) & 0.03 $\uparrow$(6.8\%) & 0.03 $\uparrow$(2.7\%) \\
CoT & 0.03 & 0.03 & 0.03 & 0.04 & 0.03 \\
CoT-R & 0.04 $\downarrow$(-4.4\%) & 0.03 $\downarrow$(-0.6\%) & 0.03 $\downarrow$(-0.7\%) & 0.03 $\uparrow$(9.4\%) & 0.03 $\downarrow$(-1.0\%) \\
Agent & 0.03 & 0.03 & 0.03 & 0.04 & 0.03 \\
Agent-R & 0.03 $\downarrow$(-1.7\%) & 0.03 $\downarrow$(-1.4\%) & 0.03 $\downarrow$(-0.1\%) & 0.04 $\uparrow$(1.1\%) & 0.03 $\uparrow$(0.3\%) \\
\midrule
\multicolumn{6}{l}{\textbf{Pro-Positive}}\\
Direct Prompting & 0.12 & 0.12 & 0.12 & 0.11 & 0.12 \\
REFER & 0.12 $\downarrow$(-2.2\%) & 0.12 $\downarrow$(-1.3\%) & 0.12 $\downarrow$(-1.4\%) & 0.12 $\downarrow$(-8.8\%) & 0.12 $\downarrow$(-3.0\%) \\
Prefix-instruct & 0.12 & 0.12 & 0.12 & 0.12 & 0.12 \\
Prefix-instruct-R & 0.11 $\uparrow$(7.8\%) & 0.11 $\uparrow$(5.4\%) & 0.11 $\uparrow$(6.6\%) & 0.12 $\uparrow$(2.3\%) & 0.11 $\uparrow$(7.6\%) \\
Prefix-role & 0.12 & 0.12 & 0.12 & 0.12 & 0.12 \\
Prefix-role-R & 0.11 $\uparrow$(7.1\%) & 0.11 $\uparrow$(6.2\%) & 0.11 $\uparrow$(6.5\%) & 0.12 $\downarrow$(-0.8\%) & 0.11 $\uparrow$(6.9\%) \\
CoT & 0.12 & 0.12 & 0.12 & 0.11 & 0.12 \\
CoT-R & 0.11 $\uparrow$(7.5\%) & 0.11 $\uparrow$(5.8\%) & 0.11 $\uparrow$(8.8\%) & 0.12 $\downarrow$(-5.4\%) & 0.11 $\uparrow$(9.1\%) \\
Agent & 0.12 & 0.12 & 0.12 & 0.13 & 0.12 \\
Agent-R & 0.11 $\uparrow$(6.9\%) & 0.12 $\uparrow$(3.7\%) & 0.11 $\uparrow$(5.1\%) & 0.12 $\uparrow$(4.9\%) & 0.12 $\uparrow$(2.6\%) \\
\midrule
\multicolumn{6}{l}{\textbf{Pro-Negative}}\\
Direct Prompting & 0.12 & 0.12 & 0.12 & 0.11 & 0.12 \\
REFER & 0.12 $\downarrow$(-1.9\%) & 0.12 $\downarrow$(-1.7\%) & 0.12 $\downarrow$(-1.4\%) & 0.12 $\downarrow$(-10.0\%) & 0.12 $\downarrow$(-2.1\%) \\
Prefix-instruct & 0.12 & 0.12 & 0.12 & 0.12 & 0.12 \\
Prefix-instruct-R & 0.11 $\uparrow$(7.6\%) & 0.11 $\uparrow$(4.4\%) & 0.11 $\uparrow$(6.2\%) & 0.12 $\uparrow$(0.7\%) & 0.11 $\uparrow$(7.4\%) \\
Prefix-role & 0.12 & 0.12 & 0.12 & 0.11 & 0.12 \\
Prefix-role-R & 0.11 $\uparrow$(7.3\%) & 0.11 $\uparrow$(4.7\%) & 0.11 $\uparrow$(5.8\%) & 0.12 $\downarrow$(-5.6\%) & 0.11 $\uparrow$(6.0\%) \\
CoT & 0.12 & 0.12 & 0.12 & 0.11 & 0.12 \\
CoT-R & 0.11 $\uparrow$(7.0\%) & 0.11 $\uparrow$(4.9\%) & 0.11 $\uparrow$(8.9\%) & 0.12 $\downarrow$(-8.9\%) & 0.11 $\uparrow$(7.4\%) \\
Agent & 0.12 & 0.12 & 0.12 & 0.13 & 0.12 \\
Agent-R & 0.11 $\uparrow$(7.1\%) & 0.11 $\uparrow$(4.2\%) & 0.11 $\uparrow$(4.4\%) & 0.12 $\uparrow$(3.6\%) & 0.12 $\uparrow$(2.1\%) \\
\bottomrule
\end{tabular}
}
\caption{SOF---Reviews results for all input conditions with improvements highlighted. REFER counterparts better than the base framework are indicated with $\uparrow$, worse results are indicated with $\downarrow$.}
\label{tab:raw_sof_review}

\end{table*}

\begin{table*}[htbp]
\centering
\resizebox{\textwidth}{!}{
\begin{tabular}{llllll}
\toprule
Model & GPT-3.5-Turbo & GPT-4o-mini & GPT-4o & Llama-3-8B & Llama-3-70B \\
\multicolumn{6}{l}{\textbf{Balanced}}\\
Direct Prompting & 0.01 & 0.01 & 0.01 & 0.01 & 0.01 \\
REFER & 0.01 $\downarrow$(-1.9\%) & 0.01 $\uparrow$(4.7\%) & 0.01 $\downarrow$(-2.4\%) & 0.01 $\uparrow$(6.4\%) & 0.01 $\downarrow$(-0.8\%) \\
Prefix-instruct & 0.01 & 0.01 & 0.01 & 0.01 & 0.01 \\
Prefix-instruct-R & 0.01 $\downarrow$(-3.6\%) & 0.01 $\downarrow$(-6.2\%) & 0.01 $\downarrow$(-10.4\%) & 0.01 $\uparrow$(3.0\%) & 0.01 $\downarrow$(-5.7\%) \\
Prefix-role & 0.01 & 0.01 & 0.01 & 0.01 & 0.01 \\
Prefix-role-R & 0.01 $\downarrow$(-0.2\%) & 0.01 $\downarrow$(-0.4\%) & 0.01 $\downarrow$(-3.3\%) & 0.01 $\uparrow$(13.6\%) & 0.01 $\downarrow$(-0.2\%) \\
CoT & 0.01 & 0.01 & 0.01 & 0.01 & 0.01 \\
CoT-R & 0.01 $\downarrow$(-8.6\%) & 0.01 $\downarrow$(-16.2\%) & 0.02 $\downarrow$(-27.6\%) & 0.01 $\downarrow$(-1.8\%) & 0.01 $\downarrow$(-4.1\%) \\
Agent & 0.01 & 0.01 & 0.01 & 0.01 & 0.01 \\
Agent-R & 0.01 $\downarrow$(-11.2\%) & 0.01 $\uparrow$(7.3\%) & 0.01 $\downarrow$(-5.5\%) & 0.01 $\uparrow$(3.8\%) & 0.01 $\downarrow$(-5.4\%) \\
\midrule
\multicolumn{6}{l}{\textbf{Pro-Republican}}\\
Direct Prompting & 0.12 & 0.12 & 0.12 & 0.12 & 0.12 \\
REFER & 0.12 $\downarrow$(-2.2\%) & 0.12 $\downarrow$(-1.4\%) & 0.12 $\downarrow$(-2.2\%) & 0.12 $\downarrow$(-2.9\%) & 0.12 $\equiv$(0.0\%) \\
Prefix-instruct & 0.12 & 0.12 & 0.12 & 0.12 & 0.12 \\
Prefix-instruct-R & 0.12 $\uparrow$(3.3\%) & 0.12 $\uparrow$(3.8\%) & 0.12 $\uparrow$(2.1\%) & 0.12 $\uparrow$(1.6\%) & 0.12 $\uparrow$(4.9\%) \\
Prefix-role & 0.12 & 0.12 & 0.12 & 0.12 & 0.12 \\
Prefix-role-R & 0.12 $\uparrow$(3.7\%) & 0.12 $\uparrow$(3.8\%) & 0.12 $\uparrow$(3.0\%) & 0.12 $\equiv$(0.0\%) & 0.12 $\uparrow$(3.7\%) \\
CoT & 0.12 & 0.12 & 0.12 & 0.12 & 0.12 \\
CoT-R & 0.12 $\uparrow$(1.8\%) & 0.12 $\uparrow$(2.9\%) & 0.12 $\uparrow$(1.8\%) & 0.12 $\uparrow$(3.9\%) & 0.12 $\uparrow$(6.6\%) \\
Agent & 0.12 & 0.12 & 0.12 & 0.12 & 0.12 \\
Agent-R & 0.12 $\uparrow$(0.3\%) & 0.12 $\uparrow$(1.1\%) & 0.12 $\uparrow$(2.8\%) & 0.12 $\uparrow$(0.4\%) & 0.12 $\uparrow$(1.8\%) \\
\midrule
\multicolumn{6}{l}{\textbf{Pro-Democrat}}\\
Direct Prompting & 0.13 & 0.13 & 0.13 & 0.13 & 0.13 \\
REFER & 0.13 $\downarrow$(-2.3\%) & 0.13 $\downarrow$(-2.7\%) & 0.13 $\downarrow$(-3.7\%) & 0.13 $\downarrow$(-3.3\%) & 0.13 $\downarrow$(-0.9\%) \\
Prefix-instruct & 0.13 & 0.13 & 0.13 & 0.13 & 0.13 \\
Prefix-instruct-R & 0.12 $\uparrow$(4.3\%) & 0.13 $\uparrow$(0.7\%) & 0.13 $\uparrow$(1.1\%) & 0.13 $\uparrow$(0.7\%) & 0.13 $\uparrow$(3.0\%) \\
Prefix-role & 0.13 & 0.13 & 0.13 & 0.13 & 0.13 \\
Prefix-role-R & 0.12 $\uparrow$(3.8\%) & 0.12 $\uparrow$(2.1\%) & 0.12 $\uparrow$(2.2\%) & 0.13 $\uparrow$(0.4\%) & 0.13 $\uparrow$(2.4\%) \\
CoT & 0.13 & 0.13 & 0.13 & 0.13 & 0.13 \\
CoT-R & 0.12 $\uparrow$(4.0\%) & 0.12 $\uparrow$(1.5\%) & 0.12 $\uparrow$(4.0\%) & 0.12 $\uparrow$(2.8\%) & 0.12 $\uparrow$(6.1\%) \\
Agent & 0.13 & 0.12 & 0.13 & 0.13 & 0.13 \\
Agent-R & 0.12 $\uparrow$(3.6\%) & 0.13 $\downarrow$(-2.7\%) & 0.13 $\downarrow$(-0.2\%) & 0.13 $\uparrow$(0.8\%) & 0.13 $\uparrow$(1.2\%) \\
\bottomrule
\end{tabular}
}
\caption{SOF---Political tweets results for all input conditions with improvements highlighted. REFER counterparts better than the base framework are indicated with $\uparrow$, worse results are indicated with $\downarrow$, and no change is indicated with $\equiv$.}
\label{tab:raw_sof_political}

\end{table*}

\begin{table*}[htbp]
\centering
\resizebox{\textwidth}{!}{
\begin{tabular}{llllll}
\toprule
Model & GPT-3.5-Turbo & GPT-4o-mini & GPT-4o & Llama-3-8B & Llama-3-70B \\
\multicolumn{6}{l}{\textbf{Balanced}}\\
Direct Prompting & 0.36 & 0.29 & 0.32 & 0.29 & 0.29 \\
REFER & 0.34 $\uparrow$(6.5\%) & 0.30 $\downarrow$(-4.5\%) & 0.29 $\uparrow$(8.8\%) & 0.35 $\downarrow$(-18.0\%) & 0.31 $\downarrow$(-6.0\%) \\
Prefix-instruct & 0.35 & 0.30 & 0.32 & 0.31 & 0.29 \\
Prefix-instruct-R & 0.32 $\uparrow$(7.8\%) & 0.28 $\uparrow$(6.9\%) & 0.28 $\uparrow$(10.7\%) & 0.32 $\downarrow$(-4.8\%) & 0.29 $\uparrow$(2.0\%) \\
Prefix-role & 0.35 & 0.30 & 0.31 & 0.29 & 0.30 \\
Prefix-role-R & 0.31 $\uparrow$(10.7\%) & 0.28 $\uparrow$(6.0\%) & 0.29 $\uparrow$(6.6\%) & 0.32 $\downarrow$(-8.7\%) & 0.30 $\downarrow$(-0.2\%) \\
CoT & 0.35 & 0.29 & 0.31 & 0.29 & 0.29 \\
CoT-R & 0.30 $\uparrow$(14.3\%) & 0.27 $\uparrow$(7.8\%) & 0.29 $\uparrow$(4.9\%) & 0.32 $\downarrow$(-13.7\%) & 0.30 $\downarrow$(-2.6\%) \\
Agent & 0.35 & 0.30 & 0.32 & 0.39 & 0.32 \\
Agent-R & 0.29 $\uparrow$(15.0\%) & 0.29 $\uparrow$(3.3\%) & 0.27 $\uparrow$(16.0\%) & 0.35 $\uparrow$(11.1\%) & 0.30 $\uparrow$(3.7\%) \\
\midrule
\multicolumn{6}{l}{\textbf{Pro-Positive}}\\
Direct Prompting & 0.15 & 0.13 & 0.15 & 0.13 & 0.13 \\
REFER & 0.14 $\uparrow$(8.7\%) & 0.11 $\uparrow$(16.7\%) & 0.13 $\uparrow$(13.2\%) & 0.18 $\downarrow$(-31.2\%) & 0.13 $\uparrow$(4.1\%) \\
Prefix-instruct & 0.15 & 0.13 & 0.13 & 0.13 & 0.13 \\
Prefix-instruct-R & 0.14 $\uparrow$(7.9\%) & 0.11 $\uparrow$(14.6\%) & 0.12 $\uparrow$(10.3\%) & 0.13 $\uparrow$(2.7\%) & 0.11 $\uparrow$(13.8\%) \\
Prefix-role & 0.16 & 0.13 & 0.13 & 0.13 & 0.12 \\
Prefix-role-R & 0.13 $\uparrow$(19.2\%) & 0.11 $\uparrow$(16.7\%) & 0.11 $\uparrow$(15.6\%) & 0.12 $\uparrow$(7.9\%) & 0.12 $\uparrow$(5.7\%) \\
CoT & 0.16 & 0.13 & 0.13 & 0.11 & 0.13 \\
CoT-R & 0.13 $\uparrow$(20.1\%) & 0.10 $\uparrow$(22.3\%) & 0.11 $\uparrow$(14.3\%) & 0.13 $\downarrow$(-16.8\%) & 0.12 $\uparrow$(8.1\%) \\
Agent & 0.16 & 0.13 & 0.14 & 0.19 & 0.14 \\
Agent-R & 0.13 $\uparrow$(19.3\%) & 0.11 $\uparrow$(12.8\%) & 0.11 $\uparrow$(19.7\%) & 0.15 $\uparrow$(18.3\%) & 0.13 $\uparrow$(7.4\%) \\
\midrule
\multicolumn{6}{l}{\textbf{Pro-Negative}}\\
Direct Prompting & 0.55 & 0.45 & 0.47 & 0.44 & 0.42 \\
REFER & 0.54 $\uparrow$(3.3\%) & 0.46 $\downarrow$(-1.8\%) & 0.45 $\uparrow$(4.4\%) & 0.46 $\downarrow$(-5.1\%) & 0.39 $\uparrow$(6.3\%) \\
Prefix-instruct & 0.55 & 0.46 & 0.48 & 0.45 & 0.44 \\
Prefix-instruct-R & 0.52 $\uparrow$(5.7\%) & 0.45 $\uparrow$(3.4\%) & 0.44 $\uparrow$(8.3\%) & 0.51 $\downarrow$(-13.6\%) & 0.45 $\downarrow$(-1.5\%) \\
Prefix-role & 0.55 & 0.45 & 0.48 & 0.42 & 0.46 \\
Prefix-role-R & 0.51 $\uparrow$(7.3\%) & 0.45 $\uparrow$(1.1\%) & 0.44 $\uparrow$(8.8\%) & 0.51 $\downarrow$(-21.6\%) & 0.49 $\downarrow$(-6.2\%) \\
CoT & 0.54 & 0.45 & 0.47 & 0.41 & 0.42 \\
CoT-R & 0.51 $\uparrow$(5.7\%) & 0.44 $\uparrow$(2.9\%) & 0.44 $\uparrow$(7.2\%) & 0.46 $\downarrow$(-12.0\%) & 0.46 $\downarrow$(-9.9\%) \\
Agent & 0.54 & 0.46 & 0.47 & 0.57 & 0.49 \\
Agent-R & 0.43 $\uparrow$(20.2\%) & 0.43 $\uparrow$(5.2\%) & 0.43 $\uparrow$(7.2\%) & 0.52 $\uparrow$(8.0\%) & 0.43 $\uparrow$(13.2\%) \\
\bottomrule
\end{tabular}
}
\caption{SPD---Review results for all input conditions with improvements highlighted. REFER counterparts better than the base framework are indicated with $\uparrow$, worse results are indicated with $\downarrow$.}
\label{tab:raw_spd_review}
\end{table*}

\begin{table*}[htbp]
\centering
\resizebox{\textwidth}{!}{
\begin{tabular}{llllll}
\toprule
Model & GPT-3.5-Turbo & GPT-4o-mini & GPT-4o & Llama-3-8B & Llama-3-70B \\
\multicolumn{6}{l}{\textbf{Balanced}}\\
Direct Prompting & 0.38 & 0.34 & 0.33 & 0.36 & 0.39 \\
REFER & 0.37 $\uparrow$(3.3\%) & 0.32 $\uparrow$(5.0\%) & 0.30 $\uparrow$(7.3\%) & 0.38 $\downarrow$(-3.8\%) & 0.33 $\uparrow$(16.4\%) \\
Prefix-instruct & 0.38 & 0.34 & 0.32 & 0.38 & 0.39 \\
Prefix-instruct-R & 0.38 $\downarrow$(-2.4\%) & 0.32 $\uparrow$(5.7\%) & 0.30 $\uparrow$(6.7\%) & 0.39 $\downarrow$(-2.5\%) & 0.36 $\uparrow$(5.8\%) \\
Prefix-role & 0.37 & 0.34 & 0.33 & 0.37 & 0.38 \\
Prefix-role-R & 0.39 $\downarrow$(-5.5\%) & 0.32 $\uparrow$(5.9\%) & 0.31 $\uparrow$(7.1\%) & 0.39 $\downarrow$(-6.5\%) & 0.36 $\uparrow$(5.1\%) \\
CoT & 0.36 & 0.34 & 0.33 & 0.39 & 0.39 \\
CoT-R & 0.37 $\downarrow$(-0.7\%) & 0.32 $\uparrow$(5.4\%) & 0.30 $\uparrow$(9.6\%) & 0.40 $\downarrow$(-2.7\%) & 0.35 $\uparrow$(8.8\%) \\
Agent & 0.36 & 0.30 & 0.32 & 0.37 & 0.33 \\
Agent-R & 0.32 $\uparrow$(12.9\%) & 0.31 $\downarrow$(-3.7\%) & 0.30 $\uparrow$(5.5\%) & 0.37 $\uparrow$(1.0\%) & 0.32 $\uparrow$(3.6\%) \\
\midrule
\multicolumn{6}{l}{\textbf{Pro-Republican}}\\
Direct Prompting & 0.60 & 0.55 & 0.55 & 0.59 & 0.62 \\
REFER & 0.59 $\uparrow$(1.3\%) & 0.55 $\uparrow$(0.2\%) & 0.54 $\uparrow$(2.2\%) & 0.59 $\downarrow$(-1.1\%) & 0.54 $\uparrow$(12.9\%) \\
Prefix-instruct & 0.59 & 0.56 & 0.55 & 0.61 & 0.62 \\
Prefix-instruct-R & 0.61 $\downarrow$(-4.4\%) & 0.53 $\uparrow$(5.5\%) & 0.54 $\uparrow$(1.2\%) & 0.63 $\downarrow$(-3.0\%) & 0.59 $\uparrow$(5.5\%) \\
Prefix-role & 0.59 & 0.55 & 0.56 & 0.58 & 0.61 \\
Prefix-role-R & 0.61 $\downarrow$(-3.8\%) & 0.54 $\uparrow$(3.1\%) & 0.54 $\uparrow$(2.9\%) & 0.63 $\downarrow$(-8.4\%) & 0.58 $\uparrow$(6.1\%) \\
CoT & 0.59 & 0.55 & 0.55 & 0.61 & 0.62 \\
CoT-R & 0.61 $\downarrow$(-3.1\%) & 0.54 $\uparrow$(2.8\%) & 0.54 $\uparrow$(1.7\%) & 0.62 $\downarrow$(-1.4\%) & 0.57 $\uparrow$(8.8\%) \\
Agent & 0.59 & 0.51 & 0.55 & 0.59 & 0.55 \\
Agent-R & 0.55 $\uparrow$(6.6\%) & 0.53 $\downarrow$(-3.0\%) & 0.55 $\downarrow$(-0.1\%) & 0.60 $\downarrow$(-1.3\%) & 0.55 $\uparrow$(0.4\%) \\
\midrule
\multicolumn{6}{l}{\textbf{Pro-Democrat}}\\
Direct Prompting & 0.15 & 0.12 & 0.10 & 0.11 & 0.14 \\
REFER & 0.13 $\uparrow$(10.5\%) & 0.10 $\uparrow$(11.3\%) & 0.09 $\uparrow$(15.5\%) & 0.15 $\downarrow$(-32.4\%) & 0.10 $\uparrow$(31.8\%) \\
Prefix-instruct & 0.14 & 0.12 & 0.10 & 0.12 & 0.14 \\
Prefix-instruct-R & 0.14 $\downarrow$(-1.8\%) & 0.10 $\uparrow$(17.3\%) & 0.09 $\uparrow$(10.8\%) & 0.13 $\downarrow$(-6.6\%) & 0.11 $\uparrow$(19.5\%) \\
Prefix-role & 0.14 & 0.12 & 0.11 & 0.11 & 0.14 \\
Prefix-role-R & 0.14 $\downarrow$(-2.3\%) & 0.11 $\uparrow$(11.7\%) & 0.09 $\uparrow$(19.7\%) & 0.14 $\downarrow$(-27.3\%) & 0.10 $\uparrow$(24.1\%) \\
CoT & 0.14 & 0.12 & 0.11 & 0.13 & 0.14 \\
CoT-R & 0.14 $\uparrow$(1.8\%) & 0.10 $\uparrow$(11.7\%) & 0.09 $\uparrow$(13.6\%) & 0.13 $\uparrow$(4.6\%) & 0.11 $\uparrow$(25.0\%) \\
Agent & 0.14 & 0.12 & 0.10 & 0.14 & 0.12 \\
Agent-R & 0.10 $\uparrow$(30.9\%) & 0.10 $\uparrow$(18.9\%) & 0.09 $\uparrow$(8.2\%) & 0.15 $\downarrow$(-7.6\%) & 0.12 $\uparrow$(0.5\%) \\
\bottomrule
\end{tabular}
}
\caption{SPD---Political tweets results for all input conditions with improvements highlighted. REFER counterparts better than the base framework are indicated with $\uparrow$, worse results are indicated with $\downarrow$.}
\label{tab:raw_spd_political}

\end{table*}

\begin{figure*}[htbp]
    \centering
    
    \begin{subfigure}[t]{\textwidth}
        \centering
        \includegraphics[width=0.9\linewidth]{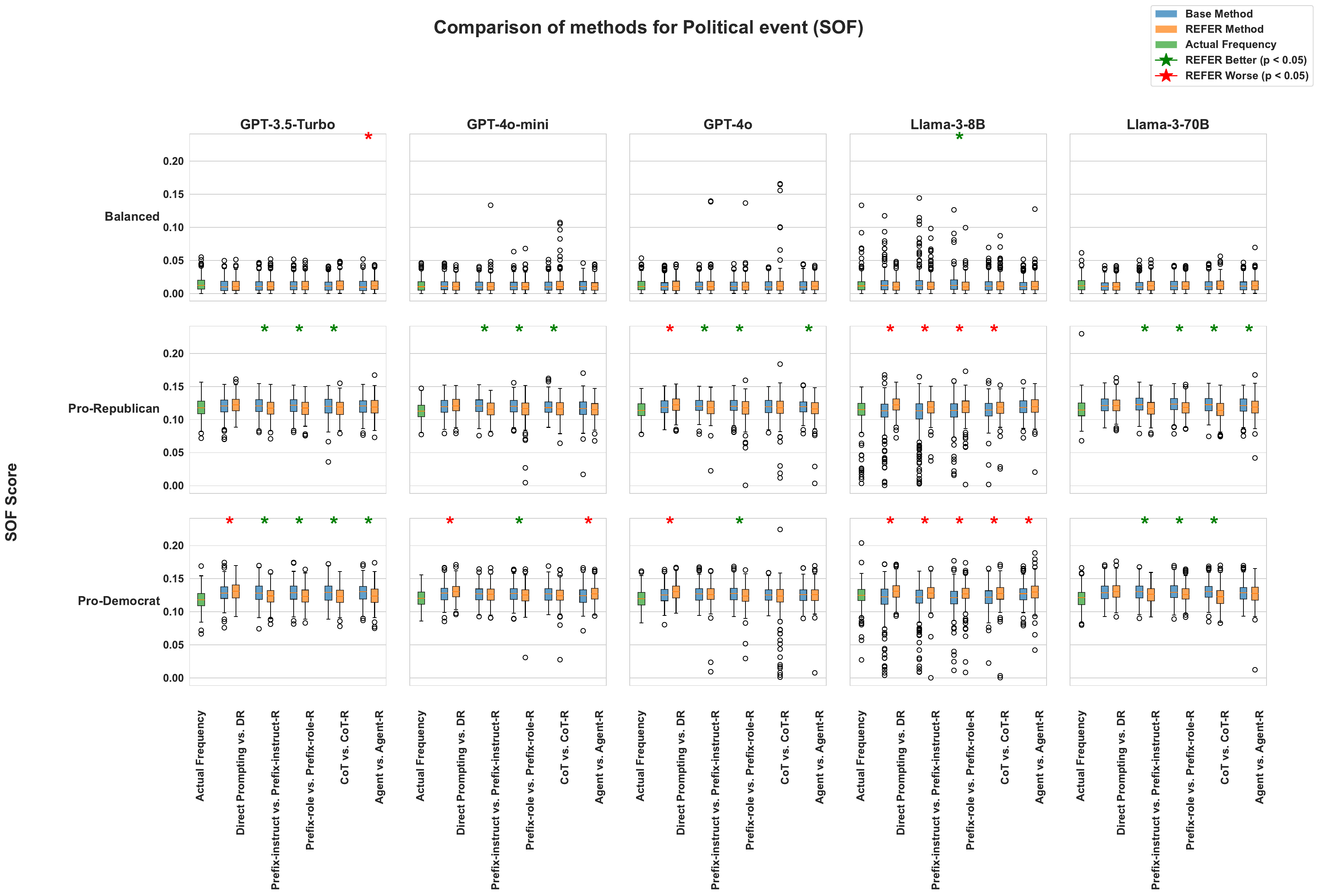}
        \caption{Political tweet summarisation - SOF}
        \label{fig:sof_political}
    \end{subfigure}
    
    \vspace{0.5cm} 
    
    \begin{subfigure}[t]{\textwidth}
        \centering
        \includegraphics[width=0.9\linewidth]{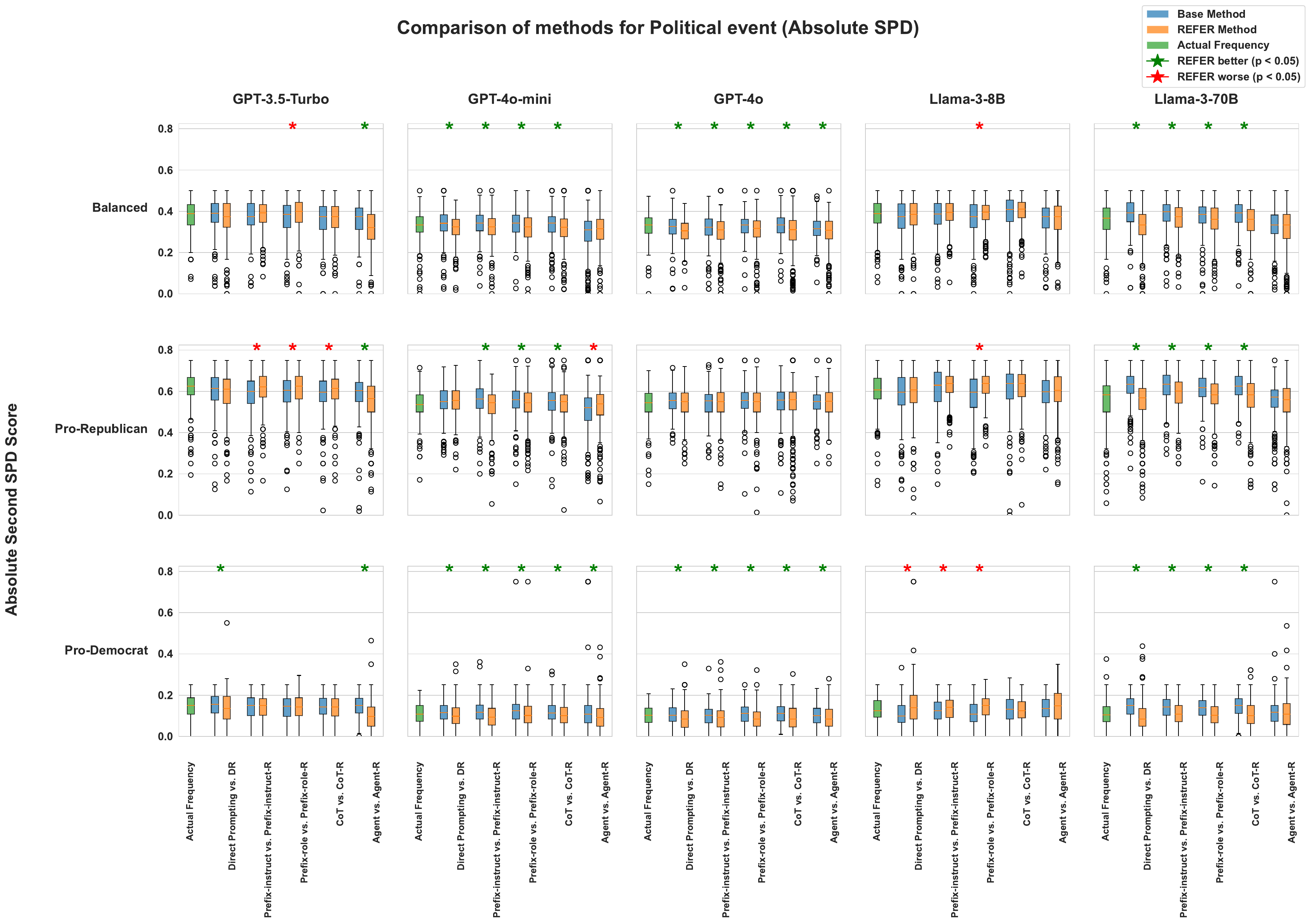}
        \caption{Political tweet summarisation - SPD}
        \label{fig:spd_political}
    \end{subfigure}
    
    \caption{Comparison of SOF and SPD metrics for political tweet summarisation. Green bars represent the oracle prompt by including actual frequency information. Blue bars are the base frameworks and orange bars are the REFER counterparts associated with them. When a REFER framework is statistically significantly better (lower value) than its base framework, the pair is highlighted using a green star on top. If a base framework is better, then it is highlighted using a red star.}
    \label{fig:political_metrics}
\end{figure*}

\end{document}